\documentclass[preprints,article,accept,moreauthors,pdftex ]{Definitions/mdpi}
\firstpage{1} 
\pdfoutput=1
\pubvolume{xx}
\issuenum{1}
\articlenumber{5}
\pubyear{2019}
\copyrightyear{2019}
\history{Received: date; Accepted: date; Published: date}

\Title{Target-Specific Action Classification for Automated Assessment of Human Motor Behavior from Video}

\Author{Behnaz Rezaei $^{1}$, Yiorgos Christakis $^{2}$, Bryan Ho $^{3}$,  Kevin Thomas $^{4}$, Kelley Erb $^{2}$, Sarah Ostadabbas $^{1}$ and Shyamal Patel $^{2*}$}

\address{%
$^{1}$ \quad Augmented Cognition Lab (ACLab), Department of Electrical and Computer Engineering, Northeastern University, Boston, MA 02115, USA; \{brezaei, ostadabbas\}@ece.neu.edu.\\
$^{2}$ \quad Digital Medicine \& Translational Imaging group, Pfizer, Cambridge, MA 02139, USA; \{Yiorgos.Christakis, MichaelKelley.Erb, Shyamal.Patel\}@pfizer.com.\\
$^{3}$ \quad Neurology Department, Tufts University School of Medicine, Boston, MA 02111, USA; bho@tuftsmedicalcenter.org.\\
$^{4}$ \quad Department of Anatomy \& Neurobiology, Boston University School of Medicine, Boston, MA 02118; kipthoma@bu.edu.}

\corres{Correspondence: Shyamal.Patel@pfizer.com.}

\usepackage[dvipdfmx]{graphicx}
\usepackage{bmpsize}
\usepackage{array}
\usepackage{times}
\usepackage{epsfig}
\usepackage{amssymb}
\usepackage[table]{xcolor}
\usepackage{array}
\usepackage{amsfonts}
\usepackage{mdwmath}
\usepackage[caption=false,font=footnotesize]{subfig}
\usepackage{balance}
\usepackage[english]{babel}
\usepackage{comment}
\newcommand{\eqnref}[1]{Equation~(\ref{eqn:#1})}
\newcommand{\figref}[1]{Fig.~\ref{fig:#1}}
\newcommand{\tblref}[1]{Table~\ref{tbl:#1}}
\newcommand{\secref}[1]{Section~\ref{sec:#1}}

\newcommand{\MethodOverview}{
\begin{figure}[!ht]
    \centering
    \includegraphics[width=0.65\textwidth, trim=2.5in 0.0in 2.5in 0.0in, clip=true]{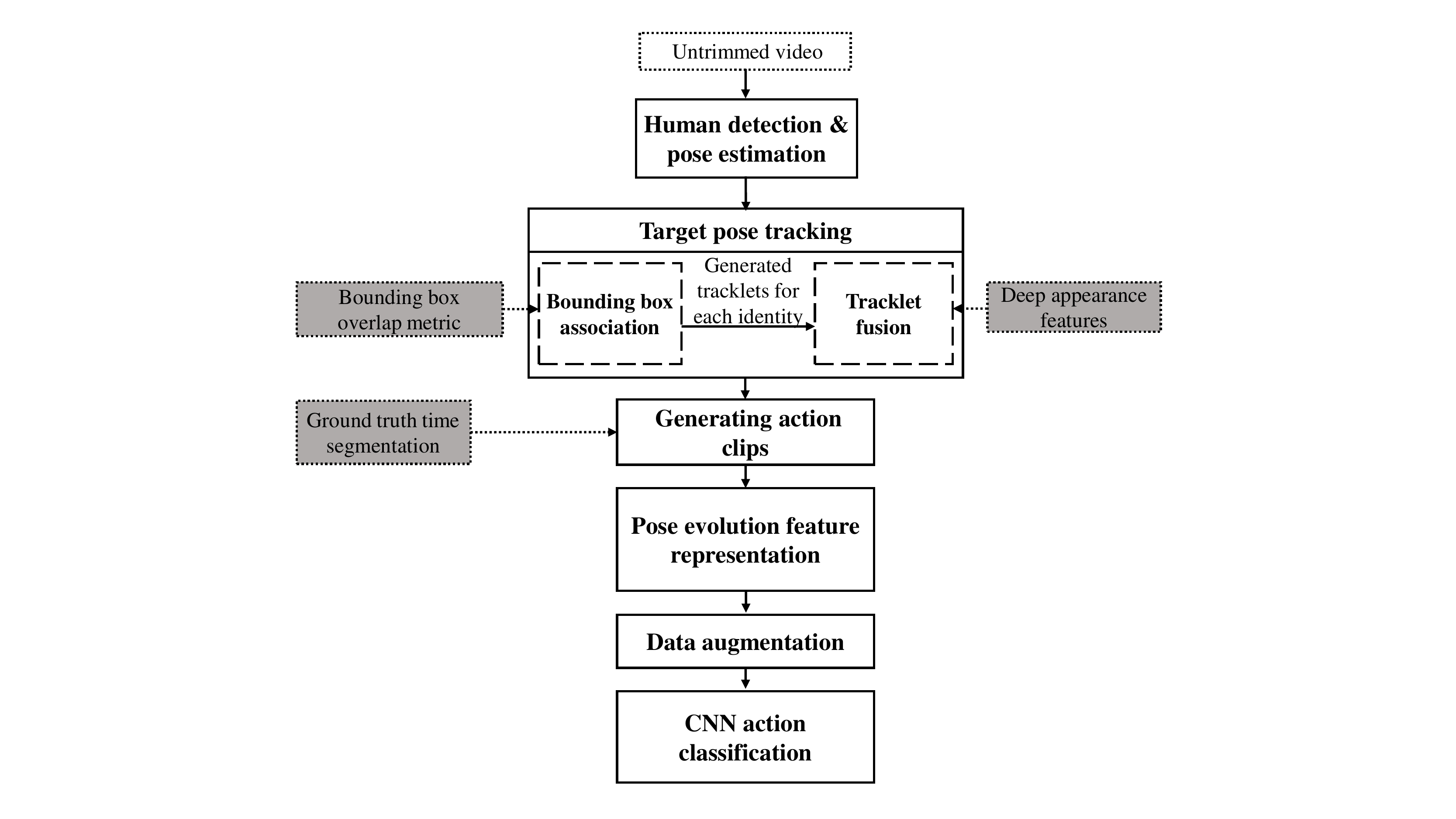}
    \caption{Overview of the proposed multi-stage method for human behavior phenotyping in untrimmed videos. At the first stage, human detection and pose estimation is applied on the recorded video. At the second stage, the regressed bounding boxes for each detected person and corresponding keypoints are used for the tracking the identities in video. Tracking is done in an incremental process incorporating both appearance and time information. Outputs of tracking the target identity along with ground-truth time segmentation are used for generating a compact representation of the target actor pose evolution in time for each action clip. Finally, the augmented pose evolution representation is fed to a CNN-based action classification network to recognize actions of interest.}
    \label{fig:Method}
\end{figure}
}
\newcommand{\PoseNet}{
\begin{figure*}[!ht]
    \centering
    \includegraphics[width=0.98\textwidth, trim=0.7in 1.6in 0.7in 1.6in, clip=true]{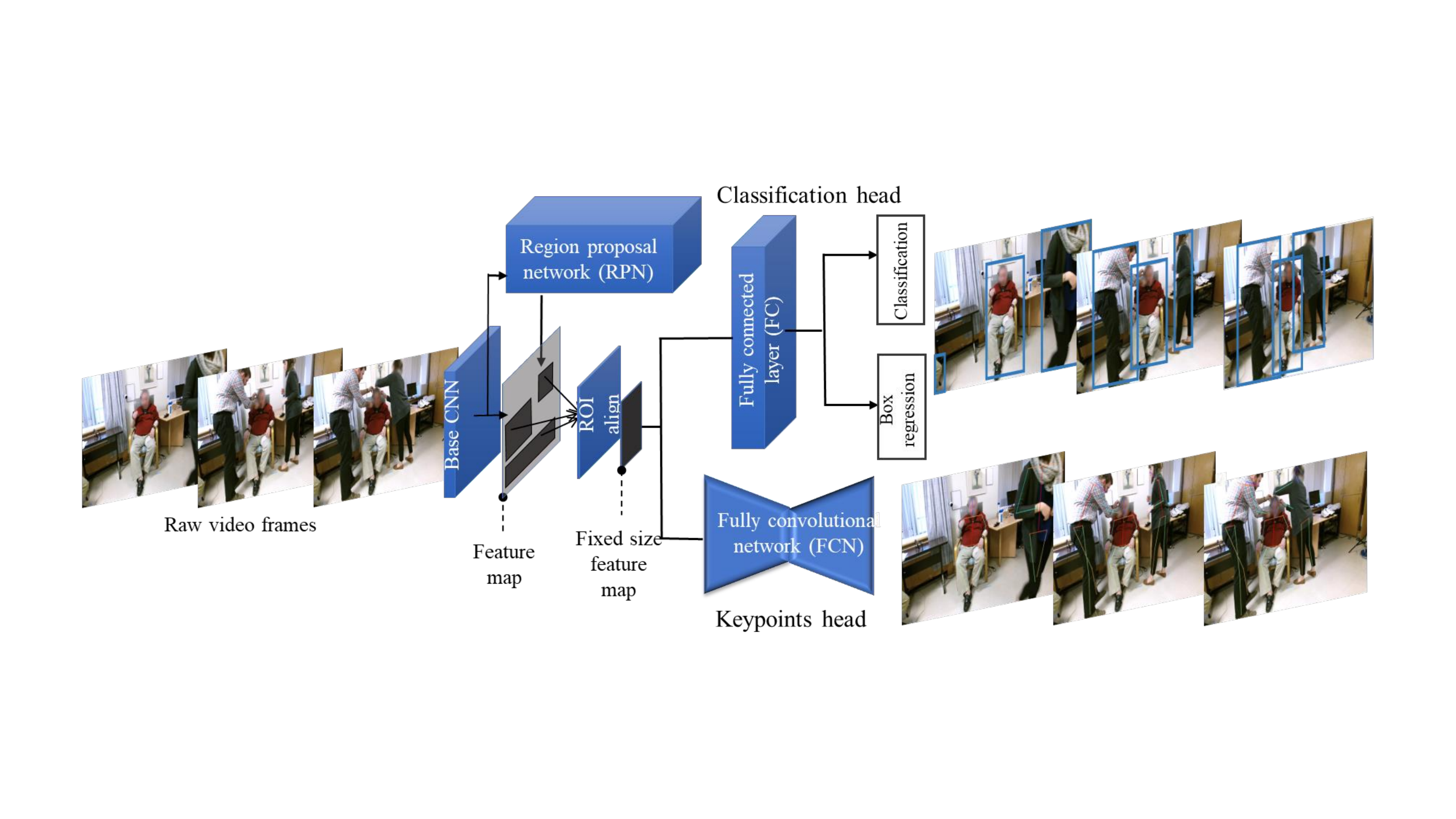}
    \caption{Architecture of the pose estimation network. Each video frame is fed separately to the base network (ResNet 101) for feature extraction. A region proposal network is applied on the output feature map to find the areas with the highest objectness probability. The fixed size features for proposed regions are then given to the classification and pose estimation heads to find the human bounding boxes and their corresponding keypoints.}
    \label{fig:Pose}
\end{figure*}
}
\newcommand{\Tracking}{
\begin{figure*}[!ht]
    \centering
    \includegraphics[width=1.0\textwidth, trim=0.1in 2.0in 0.1in 1.8in, clip=true]{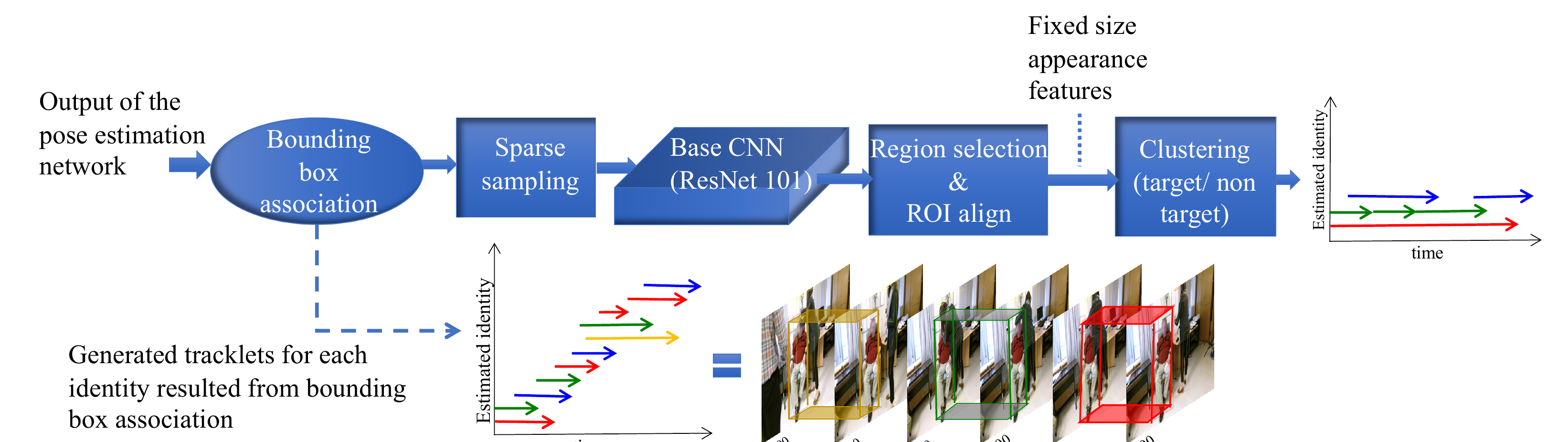}
    \caption{Hierarchical pose tracking using temporal and appearance features. Tracking starts by associating detected bounding boxes in each pair of consecutive frames using the intersection over union metric. Output of this step is a number of different tracklets for each identity. At the next step generated tracklets are pruned based on their length, and pose estimation confidence followed by sparse sampling. Finally, generated tracklets which belong to the target identity are merged according to their appearance similarity to create the endpoint track for the target human actor (best viewed in color).}
    \label{fig:Tracking}
\end{figure*}
}
\newcommand{\ActionNet}{
\begin{figure}[!ht]
    \centering
    \includegraphics[width=0.7\textwidth, trim=2.1in 1.0in 2.1in 1.0in, clip=true]{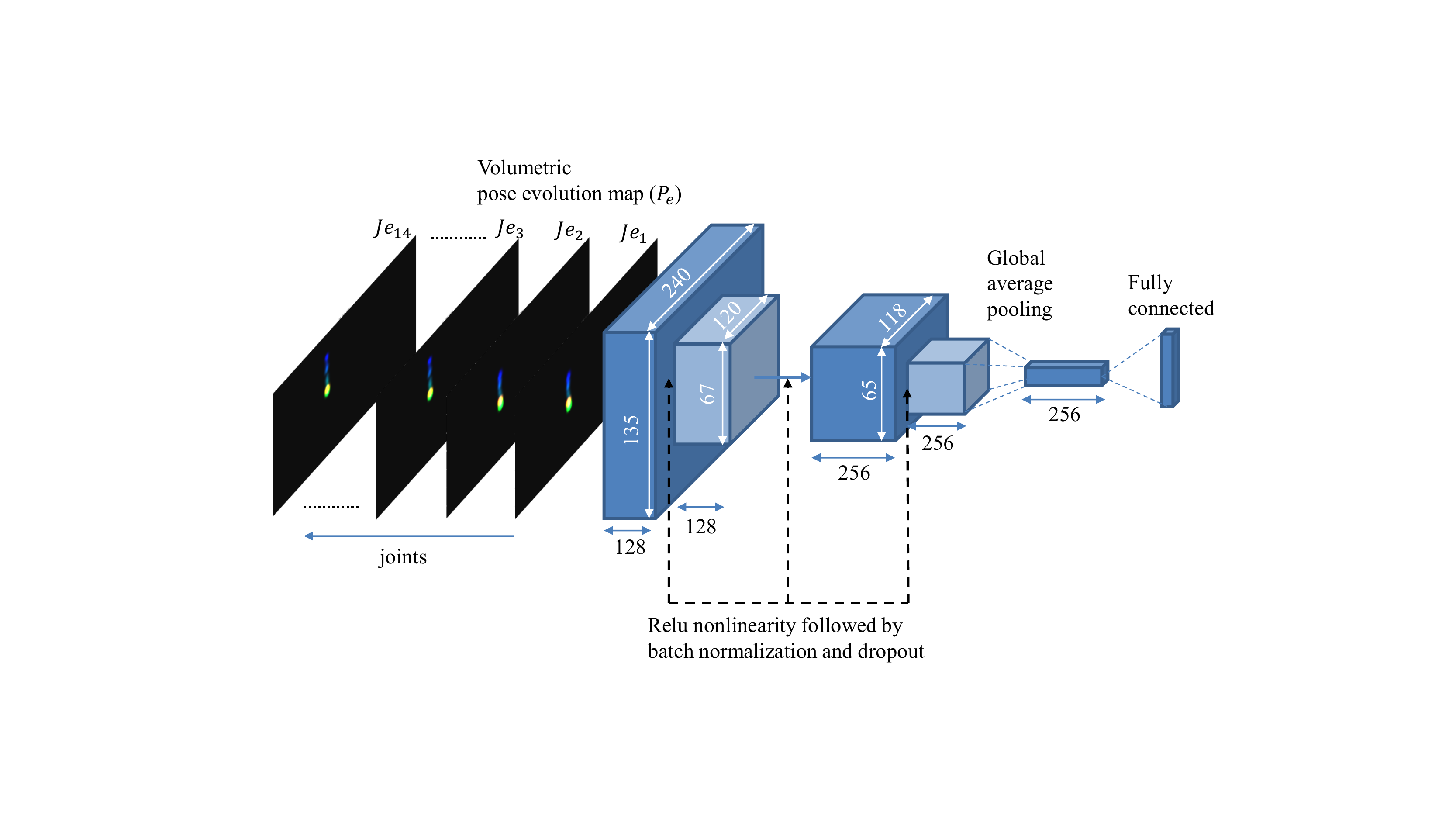}
    \caption{Architecture of the action classification network. This network takes the volumetric pose evolution map of the target human actor from a video clip as the input and classifies occurrence of an action in the video into one of the five predefined actions (best viewed in color).}
    \label{fig:ActionNet}
\end{figure}
}
\newcommand{\ColorCode}{
\begin{figure}[b]
    \centering
    \includegraphics[width=0.27\textwidth, trim=1.9in 0.9in 1.9in 0.9in, clip=true]{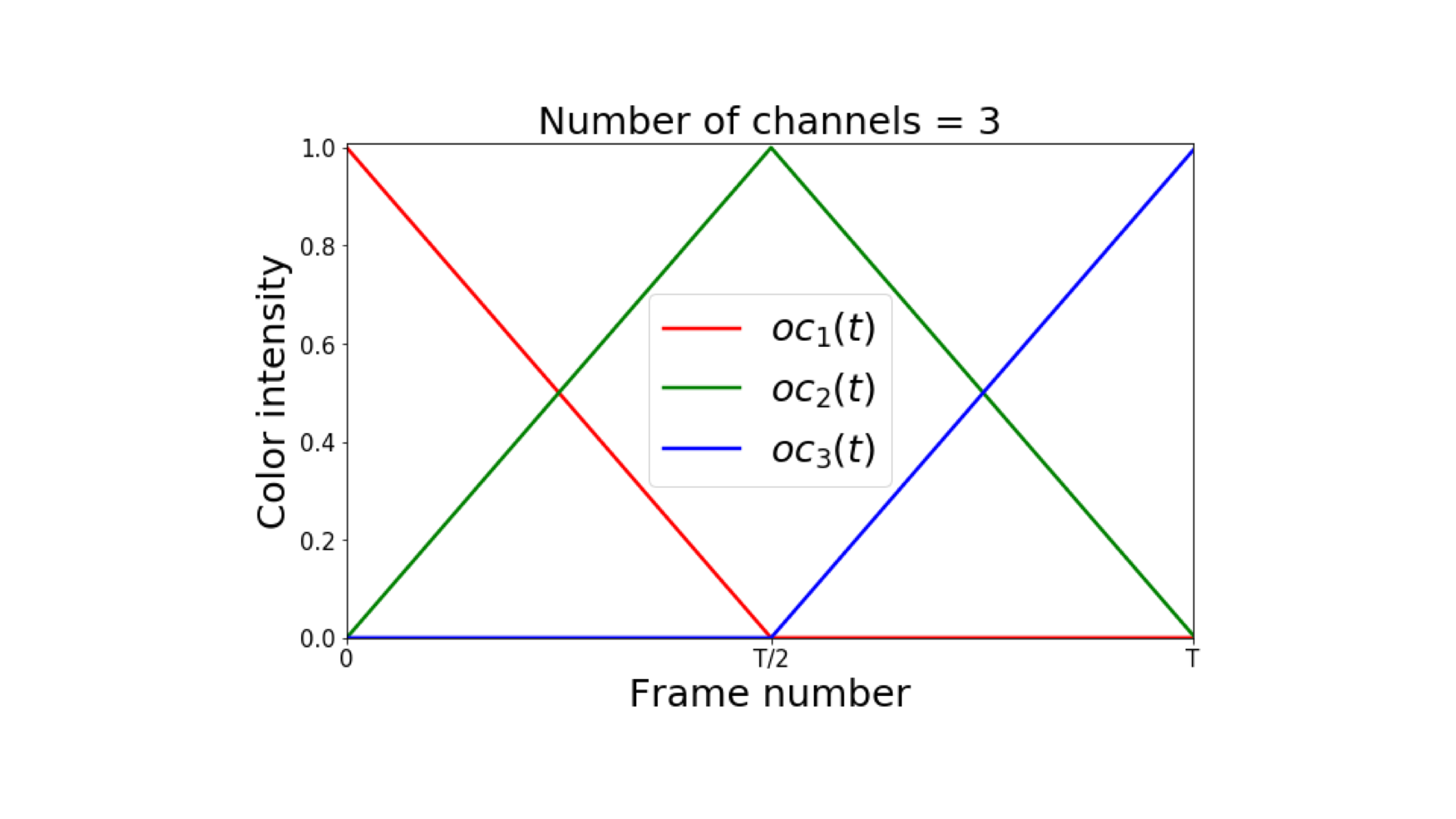}
    \caption{Demonstration of the time encoded colorization method utilized for creating body pose motion map representation. $oc_1(t)$, $oc_2(t)$, and $oc_3(t)$ show the time encoding function for each color channel.}
    \label{fig:ColorCode}
\end{figure}
}
\newcommand{\peseEvolution}{
\begin{figure}[!ht]
    \centering
    \includegraphics[width=0.8\textwidth, trim=1.0in 0.0in 1.0in 0.0in, clip=true]{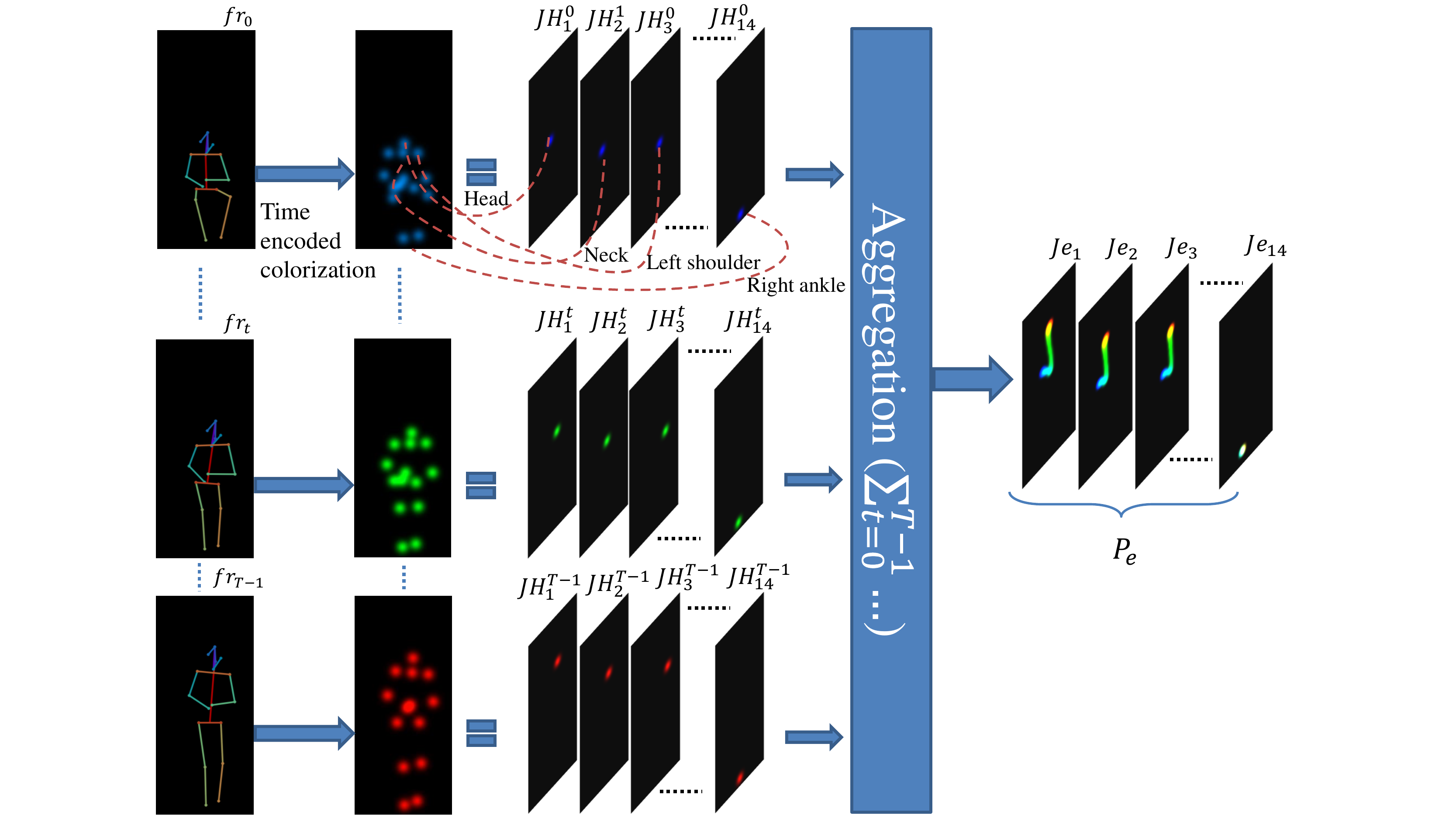}
    \caption{Illustration of the pose evolution feature representation in \figref{Method} for the sit-to-stand task. Given the estimated keypoints of the target human actor from preceding stages in the first column, colorized joint heatmaps in the second column are generated using the time encoding function represented in \figref{ColorCode}. The final pose evolution representation is generated by aggregating and normalizing the colorized joint heatmaps in time (best viewed in color).}
    \label{fig:poseEvolution}
\end{figure}
}
\newcommand{\Cmat}{
\begin{figure}[!ht]
    \centering
    \includegraphics[width=0.4\textwidth, trim=2.5in 0.0in 2.5in 0.0in, clip=true]{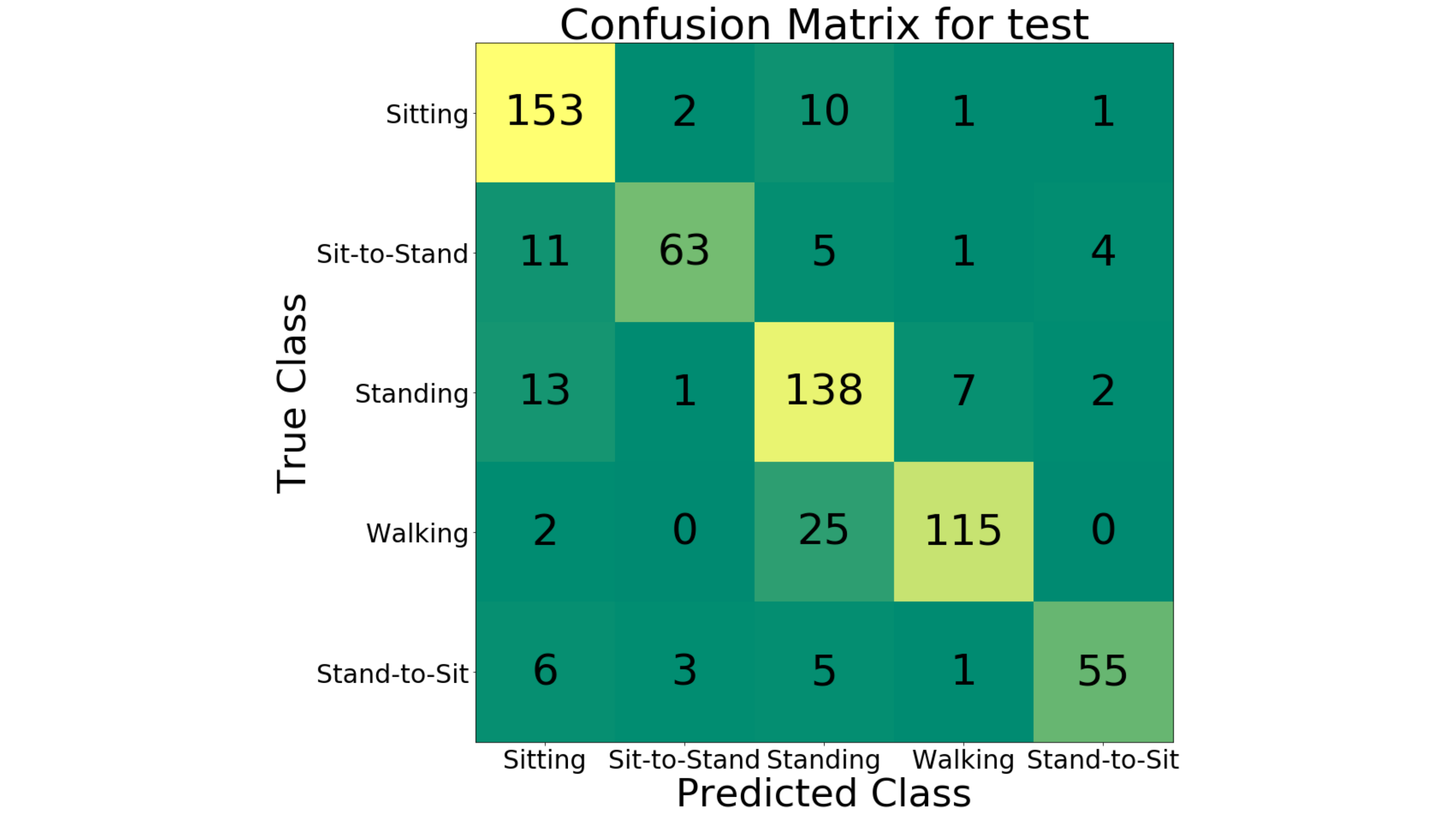}
    \caption{Confusion matrix of the action recognition network evaluated on the test dataset.}
    \label{fig:Cmat}
\end{figure}
}
\newcommand{\dataDist}{
\begin{figure}[!ht]
    \centering
    \includegraphics[width=0.7\textwidth, trim=1.1in 0.3in 1.1in 0.3in, clip=true]{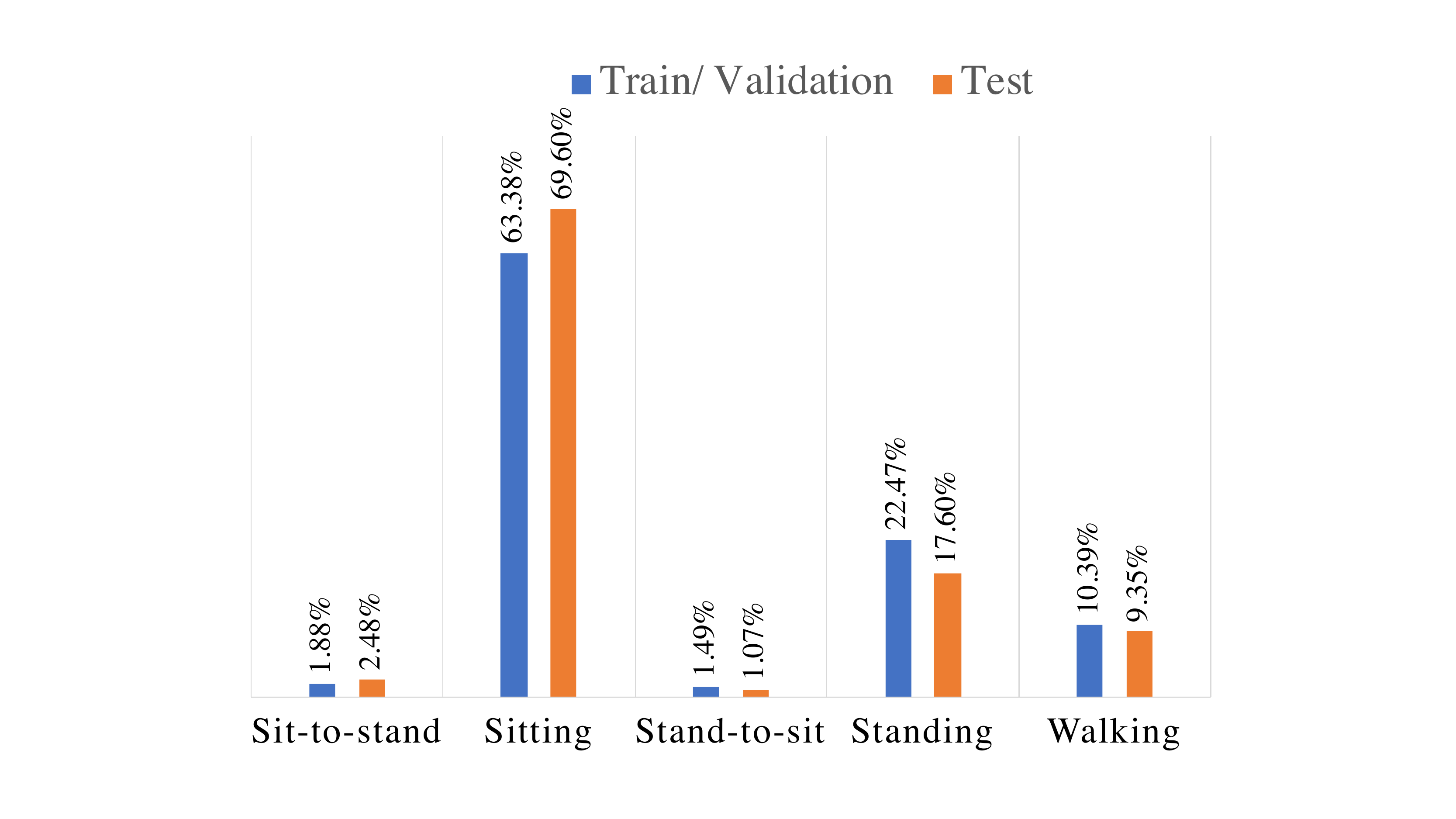}
    \caption{Distribution of the action clips based on the type of the actions for test and train/validation datasets. The distribution of the original set of action clips is highly imbalanced. }
    \label{fig:dataDist}
\end{figure}
}
\newcommand{\taxonomy}{
\begin{figure}[!ht]
    \centering
    \includegraphics[width=0.7\textwidth, trim=2.2in 2.3in 2.2in 2.3in, clip=true]{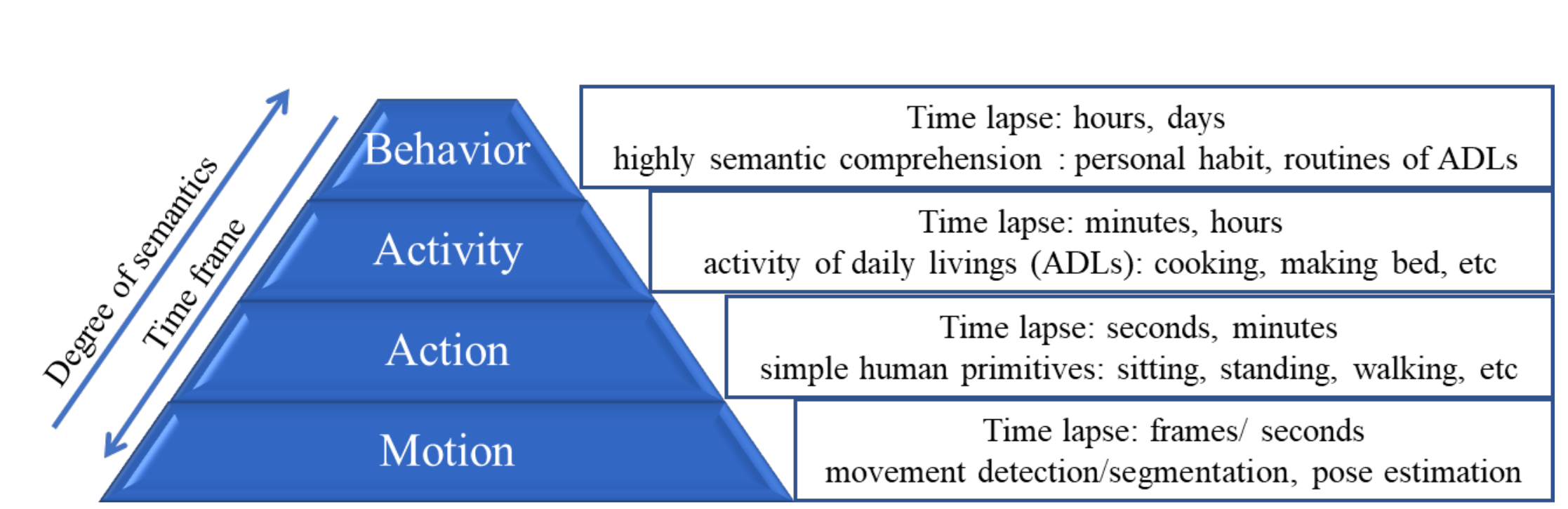}
    \caption{Taxonomy of human behaviors with different levels of semantics and complexity. Recognition of each level requires most of the underlying tasks to be recognized \cite{chaaraoui2012review}. }
    \label{fig:taxonomy}
\end{figure}
}

\newcommand{\chStat}{
\begin{figure}[!ht]
    \centering
    \includegraphics[width=0.5\textwidth, trim=2.0in 0.3in 2.0in 0.5in, clip=true]{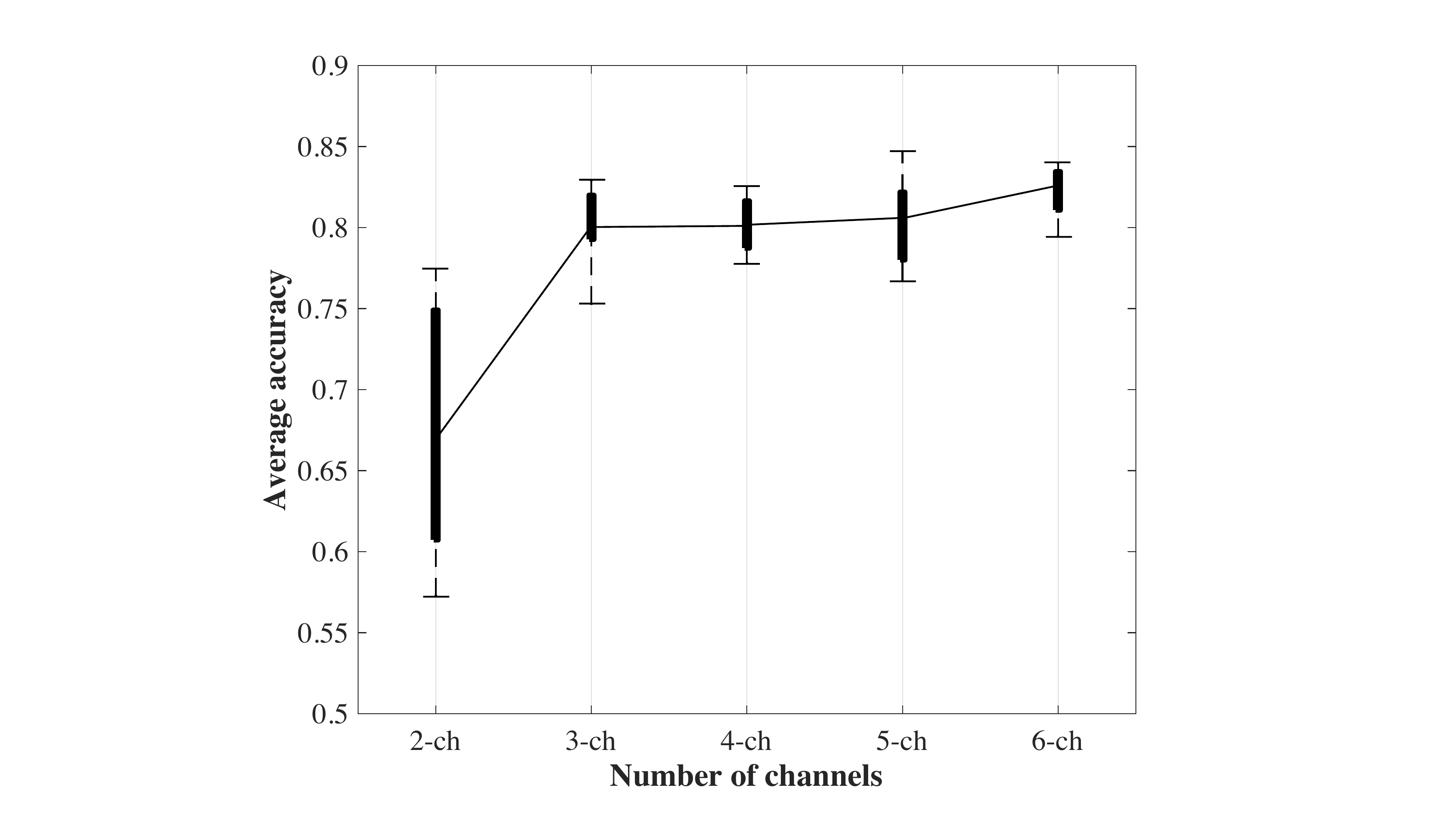}
    \caption{Average classification accuracy with respect to the number of channels of input pose evolution representations.}
    \label{fig:chStat}
\end{figure}
}

\newcommand{\missclass}{
\begin{figure}[!ht]
 \centering
  \subfloat[][\scriptsize{Standing}]{\label{fig:stand}\includegraphics[width=0.6\textwidth, trim=2.5in 2.35in 2.4in 2.5in, clip=true]{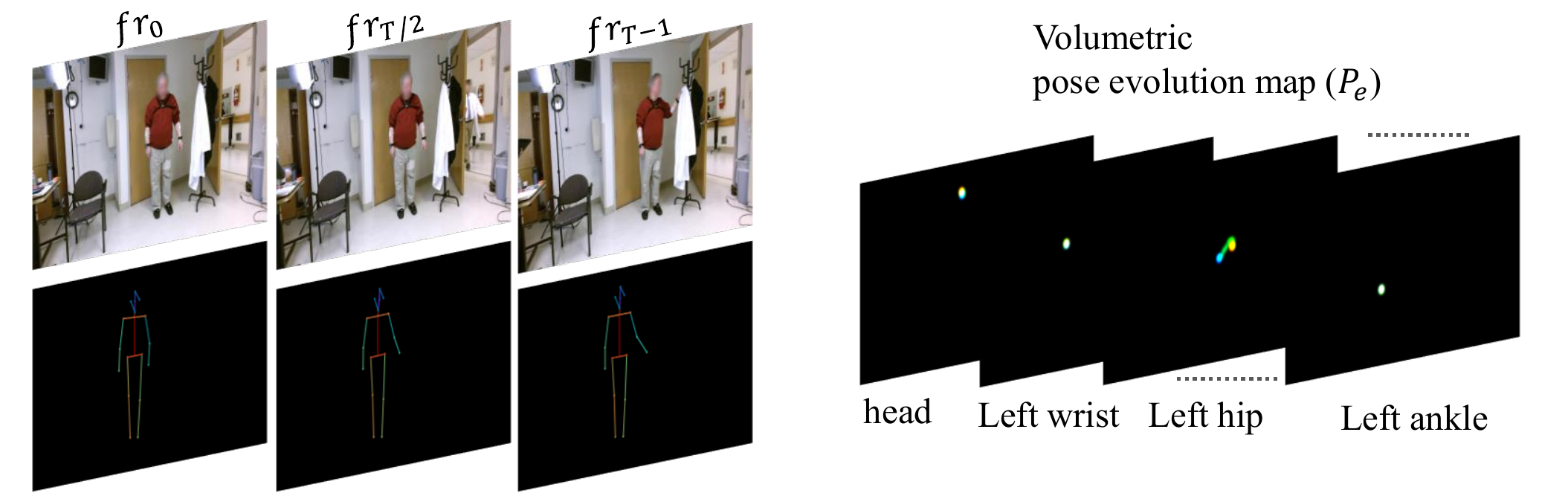}}\\
  \vspace{-.11in}
 \subfloat[][\scriptsize{Transition from standing to walking}]{\label{fig:transit}\includegraphics[width=0.6\textwidth, trim=2.5in 2.35in 2.4in 2.5in,
  clip=true]{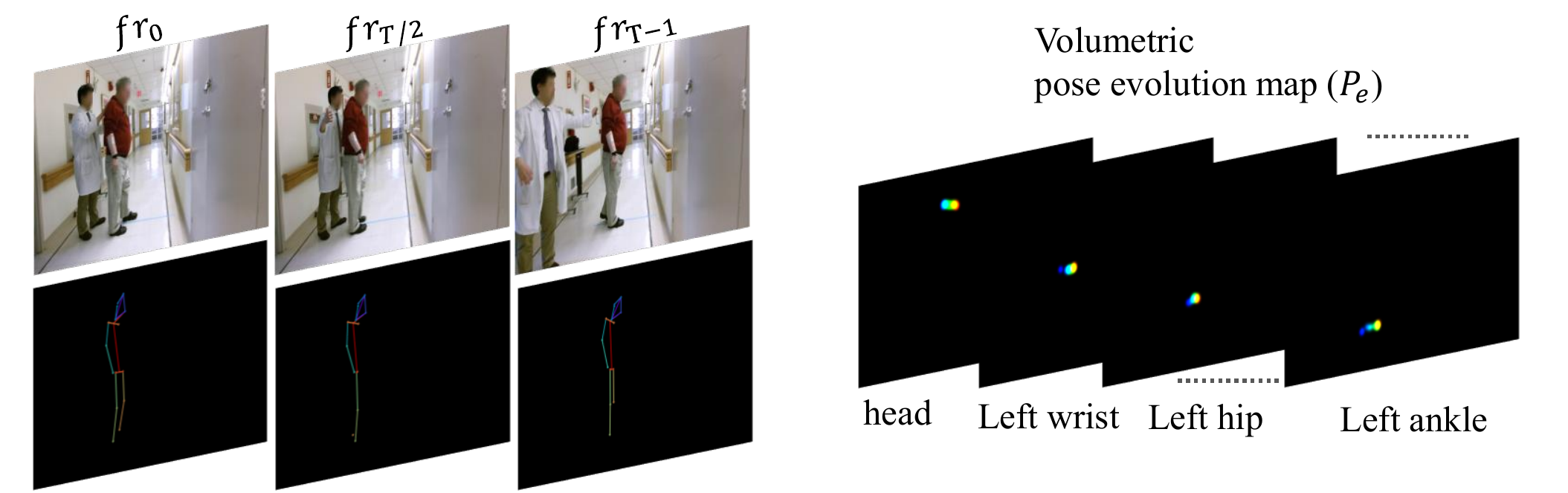}}\\
  \vspace{-.11in}
 \subfloat[][\scriptsize{Walking in frontal view}]{\label{fig:walk}\includegraphics[width=0.6\textwidth,trim=2.5in 2.35in 2.4in 2.5in,
  clip=true]{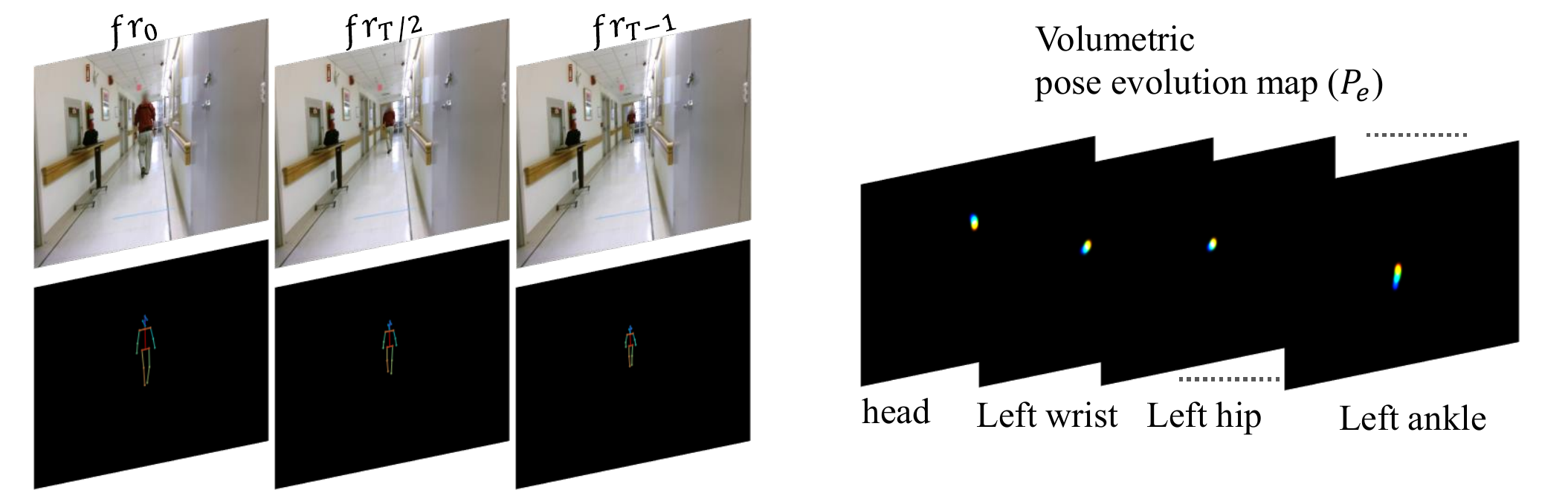}}\\
  \caption{An example of the misclassification of walking as standing. (a), (b), and (c) show the first, middle, and last frame of three action video clips along with the corresponding pose estimations and pose evolution maps. During the manual annotation process (a) was labeled as standing, whereas (b) and (c) were labeled as walking. The action classification network classifies (a) and (b) as standing because they have a very similar pose evolution map (best viewed in color and zoomed in). }
\label{fig:missclass}
\end{figure}
}
\newcommand{\accuracy}{
\begin{table}[b] 
\scriptsize
\begin{center}
 \begin{tabular}{c|c c c c c c|| c c} 

    & Sit & \shortstack{Sit-to-stand} & Stand & Walk &\shortstack{Stand-to-sit}  & \shortstack{Average accuracy \\ across activities} &  \shortstack{Mean \\accuracy} & \shortstack{Std. \\accuracy} \\ 
  \hline
  \rowcolor[gray]{0.8}
   \multicolumn{9}{c} {With long-term tracking}   \\
\hline
  Validation & 92.8 & 68.1 & 81.5 & 78.9 & 70.7 & 80.5 & 79.85  & 2.38 \\
  \hline
  Test & 91.6 & 75.0 & 85.7 & 81.0 & 78.6 & 84.0 & - & -\\
  \hline
  \rowcolor[gray]{0.8}
   \multicolumn{9}{c} {Without long-term tracking}   \\
\hline
  Validation & 90.9 &  88.1 & 91.0 & 71.8& 75.8 & 84.0 & 71.42 & 10.32 \\
  \hline
  Test     &72.6 & 63.9 & 81.6 & 51.7 & 16.3 & 63.1 & - & -
\end{tabular}
\end{center}
 \caption{\footnotesize Classification accuracy (\%) on the test and validation set in both case of using long-term tracking output and without tracking. The result shows significant drop in the performance on the test set when we do not have tracking. Standard deviation (std) of the accuracy is calculated by training the network 10 times with the same hyper-parameters but different initialization and evaluating on the validation split. }
\vspace{-.2in}
\label{tbl:accuracy}
\end{table}
}

\abstract{Objective monitoring and assessment of human motor behavior can improve the diagnosis and management of several medical conditions. Over the past decade, significant advances have been made in the use of wearable technology for continuously monitoring human motor behavior in free-living conditions. However, wearable technology remains ill-suited for applications which require monitoring and interpretation of complex motor behaviors (e.g. involving interactions with the environment). Recent advances in computer vision and deep learning have opened up new possibilities for extracting information from video recordings. In this paper, we present a hierarchical vision-based behavior phenotyping method for classification of basic human actions in video recordings performed using a single RGB camera. Our method addresses challenges associated with tracking multiple human actors and classification of actions in videos recorded in changing environments with different fields of view. We implement a cascaded pose tracker that uses temporal relationships between detections for short-term tracking and appearance based tracklet fusion for long-term tracking. Furthermore, for action classification, we use pose evolution maps derived from the cascaded pose tracker as low-dimensional and interpretable representations of the movement sequences for training a convolutional neural network. The cascaded pose tracker achieves an average accuracy of 88\% in tracking the target human actor in our video recordings, and overall system achieves average test accuracy of 84\% for target specific action classification in untrimmed video recordings.}


\begin{document}
\section{Introduction}
Clinical assessment of human motor behavior plays an important role in the diagnosis and management of medical conditions like Parkinson's Disease (PD) \cite{post2005unified}. However, such assessments can only be performed intermittently by trained clinical examiners, which limits the quantity and quality of information that can be collected to understand the impact of disease in the real-world setting. To address these limitations, significant efforts have been made to develop wearable sensing technologies that can be used for continuously monitoring various types of motor symptoms and behaviors \cite{Espay2016he, thorp2018monitoring, lara2013survey}. While data collected using wearable sensors are well suited for detecting and measuring basic movements (e.g. arm or leg movements, tremor) and actions (e.g. sitting, standing, walking), they are ill-suited when it comes to complex activities (e.g. cooking, grooming) and behaviors (e.g. personal habits, routines) - particularly if they involve the interpretation of environmental interactions (e.g. with other humans, animals, or objects). Understanding the various factors that influence physical behavior can help clinicians better understand the impact of motor and non-motor symptoms on the daily life of patients with PD \cite{vanNimwegen2011cf}.
\taxonomy

Recently, artificial intelligence (AI) assisted classification of human behavior using computer vision has received newfound attention among researchers in machine learning and pattern recognition communities for applications spanning from automatic recognition of daily life activities in smart homes to monitoring the health and safety of elderly and patients with mobility disorders in their homes/hospitals \cite{vrigkas2015review,chaaraoui2012review,chen2018robust,Li2018hs, Brattoli_2017_CVPR, song2018human, schmitt2017see}. However, in contrast to wearable devices, vision-based approaches pose a greater risk to privacy and security of an individual \cite{Chen2017cvpr}. Vision-based assessment of human behavior enables us to automate the detection and measurement of the full range of human behaviors. As illustrated in \figref{taxonomy}, the taxonomy of human behaviors can be viewed as a four-level hierarchical framework with basic movements at the bottom (e.g. movement of body segments) and complex behaviours (e.g. personal habits and routines) at the top. Automatic recognition at any level requires that actions and/or behaviors at the level below it are also recognized. For example, in order to recognize walking, we first need to assess if the pose is upright, the arms are swinging and legs are moving. At the first level (motion), recognition deals with tasks such as movement detection or background extraction/segmentation in video recordings of the target \cite{rezaei2017background, 7952497, 8456638}. These techniques try to locate the moving objects in a scene by extracting a silhouette of the object in a single frame or over a few consecutive frames. However, segmentation algorithms without any further processing provide only very basic pose estimation of the object with little to no temporal information. At the second level (action), human movements along with environmental interactions are classified in order to recognize what the target is doing over a period of seconds or minutes \cite{herath2017going}. At the third level (activity), the recognition task is focused on identifying activities as a combination of sequence of actions and environmental interactions over a period of minutes to hours. Finally, at the fourth level (behavior), sequence of activities and environmental interactions along with information about their temporal dependencies are used to recognize complex human behaviors.

\subsection{Our contributions}

Automated assessment of human behavior in multi-person video (i.e. when several people are present in the video) requires the tracking and classification of a sequence of actions performed by a target (e.g. patient). Therefore, accurate temporal tracking of the target is an essential requirement for this application, along with robust feature extraction that can be used for classifying human behaviors at different levels of complexity. In this paper, we present a hierarchical target-specific action classification method, which is illustrated as a block diagram in \figref{Method}. Detection of different actions performed by the target is done using pose evolution feature representation. We define pose evolution as a low-dimensional embedding of a sequence of posture movements that are required to perform an action (e.g. walking). In order to find the pose evolution feature representation corresponding to the target, we present a cascaded target pose tracking algorithm that receives multi-person pose estimation results from an earlier stage and tracks the target pose throughout the video. Our main contributions in this paper are: (1) development of a robust hierarchical multiple-target pose tracking method to facilitate action recognition in videos recorded in uncontrolled environments in the presence of multiple human actors; (2) introducing pose evolution, an explicit body movement representation, as complementary information to the appearance and motion cues for robust action recognition; and (3) a novel target-specific action classification architecture applied to untrimmed video recordings of patients with PD.

\subsection{Related works}
\MethodOverview
Assigning a single action label to a multi-person video clip dilutes the specificity of information and makes it less meaningful. For many real-world applications such as video-based assessment of human behavior, there is a need for person-centric action recognition, which assigns an action label to each person in a multi-person video clip. One of the challenges in person-centric action recognition is robust tracking of the target in long-term videos. Tracking is challenging because there are many sources of uncertainty, such as clutter, occlusions, target interactions, and camera motion. However, most of the research studies on human activity classification have typically dealt with videos with a single human actor or video clips with ground-truth tracking provided \cite{dawar2018data}, with the exception of few that performed human-centeric action recognition \cite{girdhar2019video,chen2018robust}. Girdhar, et al. \cite{girdhar2019video} re-purposed an action transformer network to exclude non-target human actors in the scene and aggregated spatio-temporal features around the target human actor. Chen, et al. \cite{chen2018robust} presented human activity classification using skeleton motions in videos with interference from non-target objects aimed at supporting applications in monitoring frail and elderly individuals. However, neither work provided details on how they addressed non-target filtering in their human action classification pipelines. 

Beside the importance of dealing with non-target objects in providing a well-performing real-world human action recognition system, creating robust and discriminating feature representations for each video action clip plays an important role in detecting different human activities \cite{zhang2019comprehensive}. Most of the state-of-the-art action recognition architectures process appearance and motion cues in two independent streams of information, which are fused right before the classification phase or a few stages before the classification stage in a merge and divide scheme \cite{li2018detecting, simonyan2014two}. Others have used 3D spatio-temporal convolutions to directly extract relevant spatial and temporal features \cite{zhou2018mict, tran2018closer, tran2015learning}. However, human pose cues, which can provide low-dimensional interpretations for different activities, have been overlooked in these studies. Most recently, Choutas, et al. \cite{liu2018recognizing} and Mengyuan, et al. \cite{choutas2018potion} used temporal changes of pose information with two different representations for boosting action recognition performance. In \cite{choutas2018potion}, authors claim that if there are multiple people in the scene, pose motion representation does not need the time associations of the joints to work but they did not address how their proposed method can handle multiple human actors in a video.

In general, convolutional neural network (CNN) based action recognition approaches can be divided into three different categories based on their underlying architecture: (1) spatio-temporal convolutions (3-dimensional convolutions), (2) recurrent neural networks, and (3) two stream convolutional networks. The benefit of multi-stream networks is that different modalities can be aggregated in the network to improve performance of the final action classification task. In this paper, we addressed the problem of person-centric action recognition by long-term tracking of the target human actor. In addition, our method provides a novel pose evolution representation of the target human actor rather than the common spatio-temporal features extracted from raw video frames to the classification network. It is worth mentioning that our pose-based action recognition stream can be used to augment the current multi-stream action classification networks.

The rest of the paper is organized as follows. In \secref{PoseTrack}, we describe the proposed method for tracking target human actor in untrimmed videos in order to extract appropriate pose evolution features from actions performed in a video. In \secref{action}, we describe the subsequent stages for action classification (illustrated in \figref{Method}), which include pose evolution feature representation and classification network. We present our experimental setup and performance evaluation results of the proposed method in \secref{experiment}. Finally, we discuss the results in \secref{discussion} and conclude our paper in \secref{conclusion}.
\section{Target Pose Tracking}
\label{sec:PoseTrack}
\PoseNet
Diverging from the common approach of learning spatio-temporal features from videos for action classification, pose-based action classification methods have shown promising results by providing a compact representation of human pose evolution in videos \cite{cherian2018non, choutas2018potion, liu2018recognizing, zolfaghari2017chained}. The temporal evolution of pose can be used as the only discriminating feature information for classification of human actions that involve different pose transitions (e.g. walking). This approach can further be combined with spatio-temporal features to improve the performance of context-aware action classification in the case of more complex behaviors (e.g. moving an object from one place to another).

The primary task in pose-based action classification in untrimmed videos is locating the target. This requires a robust estimation and tracking of human body poses by addressing the challenges associated with long-term videos recorded for assessment of human motor behavior. These challenges include partial to complete occlusion, change of scene, and camera motion. In this section, we propose a cascaded multi-person pose tracking method using both time and appearance features, which will be used in later steps to generate pose evolution feature representations for action classification.
\subsection{Human Pose Estimation}
\label{sec:PoseEst}

In order to extract human pose information in each video frame along with their associated bounding boxes as the first step in our system, we used a 2D version of the state-of-the-art human pose estimation method proposed in \cite{girdhar2018detect}. The pre-trained model performs efficient frame level multi-person pose estimation in videos using the Mask R-CNN network \cite{he2017mask}. This model was initialized on ImageNet \cite{deng2009imagenet} and then trained on the COCO keypoint detection task \cite{lin2014microsoft}. The Mask R-CNN network was then fine-tuned on the PoseTrack dataset \cite{andriluka2018posetrack}. The architecture of this pose estimation network is illustrated in \figref{Pose}. The network uses ResNet-101 \cite{he2016deep} as the base convolutional network for extracting image features. Extracted features are then fed to a region proposal network (RPN) trained to highlight regions that contain object candidates \cite{ren2017faster}. Candidate regions of the output feature map are all aligned to a fixed resolution via a spatial region of interest (ROI)-align operation. This operation divides feature maps that may have different sizes depending on the size of detected bounding boxes to a fixed number of sub-windows. The value for each sub-window is calculated by finding a bi-linear interpolation of four regularly sampled locations inside the sub-window. The aligned features are then fed into two heads, a \textit{classification head} responsible for person detection and bounding box regression, and a \textit{keypoint head} for estimating the human body joints defined as human pose in each detected bounding box. The outputs of this pose estimation network are seventeen keypoints associated with various body joints and a bounding box surrounding each person.

\subsection{Cascaded Pose Tracking}
\label{sec:track}
In many real-world settings where a person has to be tracked across videos recorded from different cameras located in different environments, a single tracker is unable to track the person throughout the video and all of them fail when the target leaves one environment and appears into another environment or is occluded from the camera view and then reappears in the camera's field of view \cite{gou2018systematic}. In order to address this problem of tracking people in videos recorded in multiple environments (in our case different rooms and hallways) various person re-identification methods have been proposed \cite{gou2017mom, liao2015person, ahmed2015improved}. Most of the existing re-identification (re-id) methods are supervised with the assumption of availability of large manually labeled matching identity pairs. This assumption does not hold in many practical scenarios (such as our dataset) where the model has to be generalizable for any person and providing manually labeled identity matches is not feasible. Unsupervised learning for person re-id has become important in various scenarios where the system needs to be adapted to new identities such as video surveillance applications \cite{li2018unsupervised, lv2018unsupervised}. In this work we have adapted the idea of person re-id, which is used for the matching the identities among non-overlapping cameras for tracking the target throughout the non-overlapping videos. This would address challenges such as changing environments or turning away from the camera, which can be treated as the case of re-identification across different non-overlapping cameras. In the traditional re-id problem, we typically have a gallery of images containing the images taken using different cameras for different identities. Given a probe image, the aim is to match the probe identity with the images in the gallery that belong to the same identity as the probe. In our problem of long-term tracking of the target human (patient) in videos, we have a set of tracklets and a given probe (an image of the target) and the aim is to fuse all the tracklets in the set which belong to the same identity as the probe in order to find the single track of the patient throughout the video. In contrast to the re-identification problem multiple tracklets are generated because of the failure in the tracking of the target throughout the video because of the occlusions, change of environment and abrupt camera motions.

In order to continuously track the pose of the target (i.e. the subject in our dataset) in video recordings, we propose a two step procedure based on the estimated bounding boxes and keypoints provided by the pose estimation network in \secref{PoseEst}. As illustrated in \figref{Tracking}, in the first stage (short-term tracking, \secref{Hungarian}) we use a lightweight data association approach to link the detected bounding boxes in consecutive frames into tracklets. Tracklets are a series of bounding boxes in consecutive frames associated with the same identity (person). In the next stage (long-term tracking, \secref{Fusion}), we fuse tracklets of the same identity using their learned appearance features to provide continuous tracking of the target actor across the entire video recording. The implementation details are described in \secref{Fusion}.
\\
\Tracking
\subsubsection{Short term tracking based on temporal association}
\label{sec:Hungarian}
Given the detected bounding boxes for each person in the video, we link the bounding boxes that belong to the same identity in time to create pose tracklets. Assuming that there is no abrupt movement in the video, tracklets are generated by solving a data association problem with similarity measurement defined as the intersection over union between the currently detected bounding boxes and the bounding boxes from the previous frame. Like \cite{girdhar2018detect, pirsiavash2011globally}, we formulate the task as a bipartite matching problem, and solve it using the Hungarian algorithm \cite{kuhn2005hungarian}. We initialize tracklets on the first frame and propagate the labels forward one frame at a time using the matches. Any box that does not get matched to an existing tracklet instantiates a new tracklet. This method is computationally efficient and can be adapted to any video length or any number of people. However, tracking can fail due to challenges such as abrupt camera motion, occlusions and change of scene, which can result in multiple tracklets for the same identity. For instance, as illustrated in \figref{Tracking}, short term tracking generates 3 distinct tracklets for the target in just 700 consecutive frames (~23 seconds).
\\
\subsubsection{Long term tracking using appearance based tracklet fusion}
\label{sec:Fusion}
Given the large number of tracklets generated from the previous stage (i.e. short term tracking), we fuse tracklets that belong to the same identity to generate a single long-term track for the target. As illustrated in \figref{Tracking}, in order to merge the generated tracklets belonging to the same identity throughout the video, we first apply sparse sampling by pruning the tracklets based on their length and the number of estimated keypoints, and then selecting the highest confidence bounding box from each tracklet. Finally, we merge the tracklets into a single track based on their similarity to the reference tracklet. The affinity metric between the tracklet $T_i$, and the reference tracklet $T_{ref}$, is calculated as:

\begin{align}\label{eqn:affinity}
 P_a(T_i, T_{ref}) = ||f^{t'}_i - f^{t}_{ref}||_2   
\end{align}
where $f^{t'}_i$ is feature vector of the sampled detection in tracklet $T_i$ at time $t'$, and $f^{t}_{ref}$ is feature vector of the sampled detection in reference tracklet $T_{ref}$ at time $t$. Affinity metric, $P_a(.)$ is the Euclidean distance between the above feature vectors. In order to extract deep appearance features, we feed every sampled detection of each tracklet to the base network of a Mask R-CNN (i.e. ResNet-101), which has been trained on the PoseTrack dataset for pose estimation \cite{he2017mask, he2016deep}. The extracted feature map is then aligned spatially to a fixed resolution via ROI-align operation. It is worth mentioning that we do not pay an extra computational cost for learning the features for merging the associated tracklets of the target into one track. In order to show the importance of the target tracking in the performance of the action classification network we trained the action classification network on the pose evolution maps without any tracking involved, more details are provided in \secref{actionresult}

\section{Action Classification Based on Pose Evolution Representation}
\label{sec:action}

After locating the target, providing a compact yet discriminative pose evolution feature representation for each action clip plays an essential role in recognizing different actions. To achieve this, we first provide a compact spatio-temporal representation of the target's pose evolution in \secref{potion} for each video clip inspired by PoTion pose motion representation introduced in \cite{choutas2018potion}. Then, we use the tracked pose evolution to recognize five categories of human actions: sitting, sit-to-stand, standing, walking, and stand-to-sit in \secref{network}.

\subsection{Pose Evolution Representation}
\label{sec:potion}
By using pose of the target for each frame of the video clip provided by the pose tracking stage, we create a fixed-size pose evolution representation by temporally aggregating these pose maps. Pose tracking in preceding stages gives us locations of the body joints of the target (i.e. the subject in our case) in each frame of the video clip. We first generate joint heatmaps from given keypoint positions by locating a Gaussian kernel around each keypoint. These heatmaps are gray scale images showing the probability of the estimated location for each body joint. The pose evolution representations are created based on these joint heatmaps. 

As illustrated in \figref{poseEvolution}, in order to capture the temporal evolution of pose in a video clip, after generating pose heatmaps for the target actor in a video frame, we colorize them according to their relative time in the video. In other words, each gray scale joint heatmap of dimension $H \times W$ generated for the current frame at time $t$ is transformed into a C-channel color image of $C \times H \times W$. As indicated in \eqnref{poseEv}, this transformation is done by replicating the original heatmaps C times and multiplying values of each channel with a linear function of the relative time of the current frame in the video clip. 

\peseEvolution

\begin{align} \label{eqn:poseEv}
\begin{split}
   & Je_i(j, x, y) = \frac{\sum_{t=0}^{T-1} JH_i^t(x, y) \times oc_j(t)}{max_{x, y} \sum_{t=0}^{T-1} JH_i^t(x, y) \times oc_j(t) } \\
   & \text{for } i \in \{1, 2, ..., 14\},~~ j \in \{1, ..., C \} 
\end{split}  
\end{align}
where $JH_i^t(x, y)$ designates the estimated joint heatmap for joint number $i$ of the target in a given frame number $t$. $oc_j(t)$ is the linear time encoding function for channel $j$ evaluated at time $t$. $Je_i$ is the joint evolution representation for each joint $i$. The final pose evolution representation, $Pe$ is derived by concatenating all calculated joint evolutions, as $Pe = concatenate(Je_1, Je_2, ..., Je_{14})$, where we have 14 joints given by reducing the head keypoints to one single keypoint.

In order to calculate the time encoding function for a C-channel pose evolution representation, the video clip time length $T$ is divided into $C-1$ intervals with duration $l=\frac{T}{C}$ each. For each given frame at time $t$ that sits in $k$th interval which $k=\lceil \frac{t}{T} \rceil$, $oc_j(t)$ is defined as follows:
\begin{align} \label{eqn:colorFun}
     oc_j(t) = \left \{
      \begin{tabular} {ll}
          $\frac{(-t + \frac{kT}{C-1})}{l} $,  & for $j=k$\\
          $\frac{(t - \frac{T(k-1)}{C-1})}{l}$, & for $j=k+1$ \\
          $0$, & otherwise.
      \end{tabular}
     \right.
\end{align}

\ColorCode
\figref{ColorCode} illustrates the time encoding functions that are defined based on the \eqnref{colorFun} for 3-channel colorization used in our pose evolution representation. After creating the pose evolution representations, we augment them by adding white noise to our representation to train the action classification network. 

\subsection{Classification Network}
\label{sec:network}
We trained a CNN for classifying different actions using the pose evolution representations. Since pose evolution representations are very sparse and have no contextual information of the raw video frames, the network does not need to be very deep or pre-trained to be able to classify actions. We used the network architecture illustrated in \figref{ActionNet} consisting of $4$ fully convolutional layers (FCN), and one fully connected layer (FC) as the classifier. The input of the first layer is the pose evolution representation of size $14~C \times H \times W$, where 14 is the number of body joints that are used in our feature representation. In this work, we used $C=3$ as the number of channels for encoding the time information into our feature representation. In \secref{actionresult}, we explore the effect of number of channels on the performance of the action classification network.

The action classification network includes two blocks of convolutional layers, a global average pooling layer, and a fully connected layer with a Softmax loss function as the classification layer. Each block contains two convolution layers with filter sizes of $3 \times 3 \times 128$, and $3 \times 3 \times 256$, respectively. The first convolution layer in each block is designed with a stride of 2 pixels and a second layer with a stride of 1 pixel. All convolutional layers are followed by a rectified linear unit (ReLU), batch normalization, and dropout. We investigated the performance of several variations of this architecture on action classification in \secref{experiment}. 

\ActionNet
\section{Experiments} \label{sec:experiment}
To evaluate the performance of the proposed approach, we used a real-world dataset collected in a neurology clinic. We provide an overview of the dataset in \secref{dataset}, and report on the performance of target tracking and action classification in \secref{trackResult} and \secref{actionresult} respectively.

\subsection{Dataset} \label{sec:dataset}
Our dataset consists of video recordings of 35 patients with Parkinson's disease (Age: $68.31 \pm 8.03$ [46–79] years; Sex: 23M/12F; Hoehn \& Yahr I/II/III: 2/26/7; MDS-UPDRS III: $52.86 \pm 16.03$) who participated in a clinical study to assess changes in their motor symptoms before (OFF state) and after (ON state) medication intake. Individuals with a clinical diagnosis of PD between 30-80 years old, able to recognize wearing-off periods, with Hoehn \& Yahr stage $\leq$ III and currently on L-dopa therapy were eligible to participate in this study. Exclusion criteria included the presence of other comorbidities (e.g. head injuries, psychiatric illness, cardiac disorders), recent treatment with investigational drugs, pregnant women and allergy to silicone or adhesives. The study had approval from the Tufts Medical Center and Tufts University Health Sciences Institutional Review Board (study ID: HUBB121601) and all experimental procedures were conducted at Tufts Medical Center \cite{erb2018bluesky}. All subjects provided written informed consent.

The study protocol included two visits to the clinic; subjects were randomly assigned to be in the ON (after medication intake) or OFF (before medication intake) state for the first visit, and underwent the second visit in the other state. During each study visit, patients performed a battery of motor tasks including activities of daily living (e.g. dressing, writing, drinking from a cup of water, opening a door, folding clothes) and a standard battery of clinical motor assessments from the Movement Disorder Society's Unified Parkinson's Disease Rating Scale (MDS-UPDRS) \cite{goetz2008movement} administered by a trained clinician with experience in movement disorders. Each visit lasted approximately 1 hour and most of the experimental activities were video recorded at 30 frames per second by two Microsoft Kinect\textsuperscript{TM} cameras ($1080 \times 1920$-pixel resolution), one mounted on a mobile tripod in the testing room and another on a wall mount in the adjacent hallway. In total, the dataset consists of 70 video recordings (35 subjects $\times$ 2 visits per subject). The video camera was positioned to capture a frontal view of the subject at most times. Besides the subject, there are several other people (e.g. physicians, nurses, study staff) who appear in these video recordings.

Behaviors of interest were identified within each video using structured definitions and, their start and end times annotated using human raters as described elsewhere \cite{brooks2019}. Briefly, each video recording was reviewed and key behaviors annotated by two trained raters. To maximize inter-rater agreement, each behavior had been explicitly defined to establish specific, anatomically based visual cues for annotating its start and end times. The completed annotations were reviewed for agreement by an experienced arbitrator, who identified and resolved inter-rater disagreements (e.g. different start times for a behavior). The annotated behaviors were categorized into three classes: postures (e.g. walking, sitting, standing), transitions (e.g. sit-to-stand, turning), and cued behaviors (i.e. activities of daily living and MDS-UPDRS tasks). In this manuscript, we focus on the recognition of postures (sitting, standing and walking) and transitions (sitting-to-standing and standing-to-sitting). Recognizing these activities in PD patients provide valuable context for understanding motor symptoms like tremor, bradykinesia and freezing of gait. Major challenges in recognizing activities of the target (i.e. subject) in this dataset were camera motion (when not on tripod), change of scene as the experimental activities took place in different environments (e.g. physician office, clinic hallway, etc.) and long periods occlusion (around a few minutes) due to interactions between the patient and the study staff.

\subsection{Tracking Target Human and Pose} \label{sec:trackResult}
Given that video recordings involved the presence of multiple people, we first detected all human actors along with their associated keypoints in each video frame using the multi-person pose estimation method described in \secref{PoseEst}. This pose estimation network was pre-trained on the COCO dataset and fine-tuned on the PoseTrack dataset previously \cite{lin2014microsoft,andriluka2018posetrack, girdhar2018detect}. As illustrated in \figref{Tracking}, the output of this stage is a list of the bounding boxes for human actors detected in each video frame and the estimated locations of keypoints for each person along with a confidence estimate for each keypoint.

In order to recognize activities of the target, we first locate and track the subject (i.e. PD patient) in each frame. This was accomplished by using the hierarchical tracking method described in \secref{track}. Given all detected bounding boxes across all frames from the pose estimation stage, we first generate tracklets for each identity appearing in the video via short-term tracking explained in \secref{Hungarian}. Each tracklet is a list of detected bounding boxes in consecutive frames that belong to the same identity. In order to find the final patient track for the entire video, we use the long-term tracking method described in \secref{Fusion} to remove non-target tracklets (e.g. study staff, physician, nurse) and fuse the tracklets that belong to the patient using the appearance features. There is no supervision in tracking of the patient during the video except providing a reference tracklet, which is associated to with the target (i.e. subject) in the long-term tracking step.

To evaluate the performance of our target tracking method, we first manually annotated all tracklets generated by short-term tracking and then calculated accuracy of the long-term tracking method with respect to the manually generated ground-truth. Accuracy is calculated by treating the long-term tracker as a binary classifier as it excludes non-patient tracklets and fuses tracklets belonging to the target to find a single final patient track for the entire video recording. Considering patient tracklets as the positive class and non-patient tracklets as the negative class, our tracker achieved an average classification accuracy of 88\% across 70 videos on this dataset.

\subsection{Action Classification} \label{sec:actionresult}
\dataDist
\accuracy
In the last stage of our multi-stage target-specific action classification system, we trained a CNN to recognize the following five actions of interest: sitting, standing, walking, sitting-to-standing, and standing-to-sitting. After applying the target pose tracking system illustrated in \figref{Tracking}, we segmented the resulting long-term video into action clips based on ground-truth annotations provided by human raters. Although the action clips have variable lengths (ranging from a few frames to more than 10 minutes), each video clip includes only one of the five actions of interest. As a result, we ended up with a highly imbalanced dataset. In order to create a more balanced dataset for training and evaluating the action classification network, we first excluded action clips less than 0.2 sec (too short for dynamic activities like walking) and divided the ones longer than 4 seconds into 4-second clips. Assuming that 4 seconds is long enough for most activities of interest and below 0.2 seconds (lower than 6 frames) is too short to be used for recognizing an action \cite{barrouillet2004time}. This resulted in a total of 44580 action clips extracted from video recordings of 35 subjects. We used 29 subjects (39086 action clips) for training/validation set and the remaining 6 subjects (5494 action clips) were held out for testing. As shown in \figref{dataDist}, the resulting dataset is highly imbalanced with a significant skew towards the sitting class, which can result in over-fitting issues. To address this imbalance, we randomly under-sampled the walking, sitting, and standing classes to 4000 video clips each.

To prepare input data for the action classification network, we transformed each action clip into a pose evolution representation as described in \secref{potion}. To create the pose evolution maps, we scaled the original size of each video frame ($1080 \times 1920$) by a factor of $0.125$ and chose $3$ channels to represent the input based on training time and average validation accuracy in diagnostic experiments. The training dataset was also augmented by adding Gaussian noise. In addition, we tried data augmentation techniques like random translation and flipping during our diagnostic experiments, but the classification performance degraded by about $3\%$. Therefore, we only used additive Gaussian noise to randomly selected video frames as the only type of data augmentation.

We used $90\%$ of the train/validation dataset for training the action classification network with architecture illustrated in \figref{ActionNet} and the rest for validation. Network training started with random weight initialization and we used the Adam optimizer with a base learning rate of $0.01$, a batch size of $70$ and a dropout probability of $0.3$. We experimented with several variants of the network architecture proposed in \secref{action} by increasing the number of the convolution blocks to three and changing the number of filters in each block to 64, 128, 256, and 512. Based on the performance on the validation set and training loss, \figref{ActionNet} provided the best performance while avoiding over-fitting to the training data. In addition, we investigated the impact of using a different number of channels for representing the temporal pose evolution on the performance of action classification. \figref{chStat} illustrates the accuracy of the classification network with different representations as input. We chose 3 channels for our representation as adding more channels would increase the computational cost without any significant improvement in accuracy. The trained action classification model achieved average classification accuracy of $84\%$ on the test dataset. In order to show the importance of the target tracking in the performance of the action classification network, we conducted another experiment without using any tracking on the recorded videos, the results showed an average performance drop of around $11\%$ on the test set. More details of the classification performance including per class accuracy in the test and validations phase can be found in \tblref{accuracy}.
\chStat
\Cmat
\section{Discussion} \label{sec:discussion}
 \missclass
 Real-world assessment of motor behaviour can provide valuable clinical insights for diagnosis, prognosis and prevention of disorders ranging from psychiatry to neurology \cite{Insel2017hy, Arigo2019ig}. In this paper, we propose a new approach for automated assessment of target-specific behavior from video recordings in the presence of other actors, occlusions and changes in scene. This approach relies on using temporal relationships for short-term tracking and appearance-based features for long-term tracking of the target. Short-term tracking based on temporal relationships between adjacent frames resulted in $1466 \pm 653$ tracklets per video, which were then fused by using appearance-based features for long-term tracking. Using this approach, we were able to identify the target track throughout the video recording with an accuracy of 88\% in our dataset of 70 videos belonging to 35 targets (i.e. PD patients). However, one of the limitations of our dataset was that the target's appearance did not change significantly (except for a brief period when the subject put a lab coat on to perform a task) over the duration of the recording. This is unlikely in the real-world as we expect appearance to change on a daily basis (e.g. clothing, makeup) as well as over weeks and months (e.g. age or disease related changes). Therefore, the proposed method requires further validation on a larger dataset collected during daily life and would benefit from strategies for dealing with changes in appearance.

 The second aspect of our work focused on classification of activities of daily living. Activities like sitting, standing, sit-to-stand, stand-to-sit and walking are basic elements of most of the tasks that we perform during daily life. To train the activity classification model, we used pose evolution representations to capture both temporal and spatial features associated with these activities. While this model achieved a classification accuracy of 84\%, as we can see in \figref{Cmat}, a significant source of error was the misclassification of $18\%$ (25/142) of walking as standing. This could be attributed to two factors. Firstly, video recordings of the walking activity were performed with a frontal view of the subject, which limits the ability of pose evolution representations to capture features associated with spatial displacement during walking. As a result, pose evolution representations of walking and standing look similar. This would be challenging to deal with in real-world scenarios because the camera's field of view is typically fixed. This limitation highlights the need for developing methods that are robust to changes in feature maps associated with different fields of views. Secondly, the activity transition period from standing to walking was labeled as walking during the ground-truth annotation process. As a result, when the action classification network is applied to short action clips, those containing such transitions are more likely to be misclassified as standing. Examples of the aforementioned misclassification are illustrated in \figref{missclass}. In \figref{stand} the subject takes a couple of steps to reach for a coat hanging on the rack and in \figref{transit} the subject is about to start walking from a standing position. In both cases, the ground-truth annotation was walking but the video clip was classified as standing. This is a potential limitation of a video-based action recognition approach as its performance will be dependent on factors like the camera view. In comparison, approaches using one or more wearable sensors (e.g. accelerometers and gyroscopes on the thigh, chest and ankle \cite{Attal2015dk}) are relatively robust to such problems as their measurements are performed in the body frame of reference, which results in high classification accuracy (>$95\%$) across a range of daily activities.
 
 Another source of error which impacts overall performance is the error propagated from pose tracking (\textasciitilde12\%) and pose estimation stages. Pose estimation error can be tolerated to some degree by aggregating the colorized joint heatmaps in the pose evolution feature representation. However, since the output of the pose tracking is directly used for generating pose evolution maps, any error in tracking the patient throughout the video would negatively impact the action classification performance. One approach for tolerating the error from pose tracking stage is to incorporate raw RGB frames as the second stream of information for action classification and using attention maps based on the tracking outcome rather than excluding non-target persons from the input representations.
 
 Vision-based monitoring tools have the distinct advantage of being transparent to the target, which would help with issues of compliance associated with the use of wearable devices. Also, unlike wearable devices, vision-based approaches can capture contextual information, which is necessary for understanding behavior at higher level. However, this also comes at an increased risk to privacy for the target as well as other people in the environment. The proposed approach can potentially mitigate this concern by limiting monitoring to the target (e.g. patient) and transforming data at the source into sparse feature maps (i.e. pose evolution representations).
\section{Conclusion and Future Work} \label{sec:conclusion}
In this paper, we have presented an AI assisted method for automatic assessment of human motor behavior from video recorded using a single RGB camera. Results demonstrate that the multi-stage method, which includes pose estimation, target tracking and action classification, provides accurate target-specific classification of activities in the presence of other human actors and is robust to changing environments. The work presented herein focused on the classification of basic postures (sitting, standing and walking) and transitions (sitting-to-standing and standing-to-sitting), which commonly occur during the performance of many daily activities and are relevant to understanding the impact of diseases like Parkinson's disease and stroke on the functional ability of patients. This has laid the foundation for future research efforts that will be directed towards detecting and quantifying clinically meaningful information like detection of emergency events (e.g. falls, seizures) and assessment of symptom severity (e.g. gait impairments, tremor) in patients with various mobility limiting conditions. Lastly, the code and models developed during this work are being made available for the benefit of the broader research community.

\authorcontributions{Conceptualization, Behnaz Rezaei, Sarah Ostadabbas and Shyamal Patel; Data curation, Behnaz Rezaei, Yiorgos Christakis, Bryan Ho, Kevin Thomas and Kelley Erb; Formal analysis, Behnaz Rezaei; Investigation, Bryan Ho and Kelley Erb; Methodology, Behnaz Rezaei, Sarah Ostadabbas and Shyamal Patel; Project administration, Kelley Erb; Resources, Bryan Ho; Software, Behnaz Rezaei; Supervision, Sarah Ostadabbas and Shyamal Patel; Writing – original draft, Behnaz Rezaei and Shyamal Patel; Writing – review \& editing, Behnaz Rezaei, Yiorgos Christakis, Bryan Ho, Kevin Thomas, Kelley Erb, Sarah Ostadabbas and Shyamal Patel.}

\funding{Funding for this work was provided by Pfizer, Inc.}

\acknowledgments{The authors would like to acknowledge the BlueSky project team for generating the data that made this work possible. Specifically, we would like to acknowledge Hao Zhang, Steve Amato, Vesper Ramos, Paul Wacnik and Tairmae Kangarloo for their contributions to study design and data collection.}

\conflictsofinterest{S.P., Y.C. and M.K.E are employees of Pfizer, Inc. The remaining authors declare no conflict of interest}



\bibliographystyle{Definitions/mdpi}
\reftitle{References}
\bibliography{Ref}


\end{document}